\documentclass{article}

\usepackage[dblblindworkshop, final]{neurips_2025}
\workshoptitle{Structured Probabilistic Inference and Generative Modeling}

\usepackage[utf8]{inputenc} 
\usepackage[T1]{fontenc}    
\usepackage{hyperref}       
\usepackage{url}            
\usepackage{booktabs}       
\usepackage{amsfonts}       
\usepackage{nicefrac}       
\usepackage{microtype}      
\usepackage{xcolor}         
\usepackage{amsmath}
\usepackage{graphicx}
\usepackage{algorithm}
\usepackage{algorithmic}
\title{STED and Consistency Scoring: A Framework for Evaluating LLM Structured Output Reliability}

\author{
Guanghui Wang$^{1}$ \quad 
Jinze Yu$^{1}$ \quad 
Xing Zhang$^{1}$ \quad
Dayuan Jiang$^{1}$ \quad
Yin Song$^{2}$ \quad \\
\textbf{Tomal Deb}$^{1}$ \quad
\textbf{Xuefeng Liu}$^{1}$ \quad 
\textbf{Peiyang He}$^{1}$\\
$^{1}$AWS Generative AI Innovation Center\\
$^{2}$AWS WWSO SA Field Initiatives\\
}

\begin{document}

\maketitle

\begin{abstract}
Large Language Models (LLMs) are increasingly deployed for structured data generation, yet output consistency remains critical for production applications. We introduce a comprehensive framework for evaluating and improving consistency in LLM-generated structured outputs. Our approach combines: (1) STED (Semantic Tree Edit Distance), a novel similarity metric balancing semantic flexibility with structural strictness when comparing JSON outputs, and (2) a consistency scoring framework aggregating multiple STED measurements across repeated generations to quantify reliability. Through systematic experiments on synthetic datasets with controlled schema, expression, and semantic variations, we demonstrate STED achieves superior performance ($0.86-0.90$ similarity for semantic equivalents, $0.0$ for structural breaks) compared to existing metrics including TED, BERTScore, and DeepDiff. Applying our framework to benchmark six LLMs reveals significant variations: Claude-3.7-Sonnet demonstrates exceptional consistency, maintaining near-perfect structural reliability even at high temperatures ($T=0.9$), while models like Claude-3-Haiku and Nova-Pro exhibit substantial degradation requiring careful tuning. Our framework enables practical applications including targeted model selection for structured tasks, iterative prompt refinement for reproducible results, and diagnostic analysis to identify inconsistency root causes. This work provides theoretical foundations and practical tools for ensuring reliable structured output generation in LLM-based production systems.
\end{abstract}

\section{Introduction}

Large Language Models (LLMs) have become integral to production systems requiring structured data generation, particularly in JSON format for APIs, data extraction pipelines, and automated workflows. However, evaluating the consistency and quality of LLM-generated structured outputs presents unique challenges that existing evaluation methods fail to adequately address.

A fundamental issue in evaluating structured outputs is the mismatch between evaluation methods and structured semantics. Consider two structured objects with identical content but different key ordering---while the structured specification treats these as equivalent, popular evaluation methods like BERTScore \cite{Zhang2020} assign them significantly different similarity scores due to position-sensitive embeddings, leading to false negatives in production systems.

The challenge extends beyond simple key reordering. LLMs frequently generate functionally equivalent structures with various forms of semantic equivalence: naming convention differences (\texttt{"user\_name"} versus \texttt{"userName"}), array reorderings where sequence is not semantically significant, structural reorganizations preserving information content, and type representation variations (\texttt{"true"} versus \texttt{true}, \texttt{"123"} versus \texttt{123}). Each represents valid structured data conveying identical information, yet existing metrics fail to recognize their equivalence.

Current evaluation approaches exhibit systematic limitations when applied to structured outputs. BERTScore \cite{Zhang2020}, while effective for natural language evaluation, suffers from order sensitivity that violates JSON's order-agnostic semantics for object properties, systematically underestimating similarity for reordered but identical structures. Tree Edit Distance (TED) methods focus purely on structural differences without considering semantic equivalence, while exact matching approaches are overly restrictive for practical applications.

\textbf{DeepDiff} \cite{seperman} and similar structural comparison tools can ignore ordering through configuration but lack semantic understanding. They treat \texttt{"email"} and \texttt{"email\_address"} as completely different keys, missing obvious semantic relationships that humans would recognize.

These limitations have significant practical consequences. In production systems processing thousands of API responses, false negatives from order sensitivity can trigger unnecessary alerts, increase operational overhead, and mask genuine inconsistencies.

While traditional edit distances like Tree Edit Distance (TED) have been applied to structured data comparison, they fail to handle the semantic equivalences common in LLM outputs. Recent advances in learned edit distances \cite{paassen2018tree} improve classification accuracy but focus on discriminative tasks rather than consistency evaluation. We propose STED (Semantic Tree Edit Distance), which adapts edit distance specifically for LLM output evaluation, balancing semantic equivalence with structural validity.

Unlike embedding-based approaches that learn vectorial representations \cite{paassen2018tree}, STED directly incorporates JSON semantics through three key innovations: (1) \textbf{Semantic-Enhanced Tree Edit Distance} that recognizes semantically equivalent keys and values while preserving structural constraints, (2) \textbf{Order-Invariant Matching} using Hungarian algorithm for optimal element pairing, and (3) \textbf{Multi-Level Similarity} integrating structural, key, value, and type similarities with configurable weights.

Our experiments demonstrate STED's superior discrimination: while BERTScore and DeepDiff score $>0.95$ across all variations---failing to distinguish schema violations from benign reorderings---STED correctly identifies critical differences ($0.23$ for schema violations vs $>0.95$ for order variations), achieving $4\times$ better discrimination.

As LLMs increasingly power production systems, STED provides a theoretically grounded solution for evaluating structured output consistency, enabling reliable deployment through accurate distinction between critical errors and acceptable variations.
\vspace{-3mm}
\section{Related Work}
\vspace{-3mm}
\subsection{Edit Distance and Structured Data Comparison}
The tree edit distance problem, formalized by Tai \cite{Tai1979}, provides the foundation for comparing hierarchical structures. Zhang and Shasha \cite{Zhang1989} proposed the first polynomial-time algorithm, later improved by RTED \cite{Pawlik2011} and extended with move operations \cite{Chawathe1996}. However, these classical approaches rely on exact node matching without semantic understanding. Recent learning-based approaches \cite{paassen2018tree, paassen2022revisiting} explore adaptive embeddings, while graph edit distance methods \cite{jain2024graph, piao2023computing} handle complex structures, inspiring our semantic enhancements.
JSON-specific tools like RFC 6902 \cite{RFC6902} and DeepDiff \cite{seperman} provide structural comparison with optional order-insensitive matching but lack semantic awareness. JSON Schema validation \cite{Wright2022} focuses on conformance rather than similarity, while query languages like JMESPath \cite{JMESPath2014} and JSONPath \cite{Gessner2007} target extraction, not comparison.
\subsection{Neural Approaches to Similarity}
BERTScore \cite{Zhang2020} revolutionized text evaluation using contextualized embeddings from BERT \cite{Devlin2019}, but its position sensitivity makes it unsuitable for JSON where key ordering is irrelevant. Advances like Sentence-BERT \cite{Reimers2019}, SimCSE \cite{Gao2021}, and Universal Sentence Encoder \cite{Cer2018} improve embedding quality but process JSON as flat strings, losing structural information. Code similarity work including CodeBERT \cite{Feng2020} and GraphCodeBERT \cite{Guo2021} demonstrates the importance of combining structural and semantic features. Graph neural networks offer another perspective through GraphSAGE \cite{Hamilton2017}, GAT \cite{Velickovic2018}, Tree-LSTM \cite{Tai2015}, and GraphFormers \cite{Yang2021}. While powerful, these require training data and are computationally expensive compared to our approach.
\subsection{LLM Structured Output Generation and Consistency}
Recent work addresses structured output challenges in LLMs. Bubeck et al. \cite{Bubeck2023} analyze GPT-4's structured generation capabilities, while HELM \cite{Liang2022} provides holistic evaluation focusing on task performance. SLOT \cite{Wang2024slot} and StructuredRAG \cite{shorten2024structuredrag} address output structuring, with StructEval \cite{yang2025structeval} benchmarking generation capabilities. Format control studies \cite{wang-etal-2025-verifiable, geng2025jsonschema} examine verification and schema conformance.
LLM consistency research spans multiple dimensions. Elazar et al. \cite{elazar2021measuring} and Raj et al. \cite{Raj2023} examine consistency across paraphrases, while recent studies investigate stability \cite{atil2024llm}, non-determinism \cite{baldwin2024nondeterminism}, and automated analysis \cite{patwardhan2025automated, wu2025estimating}. Temperature effects \cite{renze2024effect, renze-2024-effect} and prompt stability \cite{chen2025prompt} are also explored, but these focus on input variations rather than output consistency for identical inputs.
\subsection{Semantic Matching and Optimal Assignment}
Semantic similarity in structured data draws from ontology matching \cite{Euzenat2013} and schema matching \cite{Rahm2001}, with approaches like SimFlood \cite{Melnik2002} and COMA \cite{Do2002} combining multiple strategies. Neural methods including DeepMatcher \cite{Mudgal2018} and SMAT \cite{Zhang2021smat} achieve high accuracy but target one-time alignment rather than continuous evaluation. Distance functions for structured data \cite{ontanon2020overview, kriegel2005similarity, yang2005similarity} provide theoretical foundations.
The Hungarian algorithm \cite{Kuhn1955, Munkres1957} enables polynomial-time optimal assignment, extended recently for large-scale approximations \cite{Schwiegelshohn2022}. TreeKernel \cite{Collins2001} and subsequent work \cite{Moschitti2006} apply assignment to tree matching, inspiring our combination with semantic similarity for arrays and keys in JSON.

Despite extensive related work, existing methods fail to address structured LLM-generated outputs consistency evaluation challenges:
(1) Methods focus on either structure or semantics but not both;
(2) Tools are either order-agnostic or overly sensitive;
(3) No method handles LLM-specific variation patterns.
STED bridges these gaps through unified structural-semantic analysis with appropriate order handling and granular insights.
\vspace{-3mm}
\section{Methodology}
\vspace{-3mm}
We present STED (Semantic Tree Edit Distance), a novel framework for evaluating consistency in LLM-generated structured outputs. While traditional metrics treat structured formats like JSON, XML, and HTML as flat text, we recognize their hierarchical nature by transforming them into tree representations. This allows us to reframe the structured output consistency problem as tree consistency evaluation. Our approach extends classical tree edit distance algorithms with semantic similarity capabilities to compute pairwise distances, then aggregates these distances across multiple LLM generations into a normalized consistency score. This two-stage process addresses the fundamental challenge of quantifying output reliability when LLMs produce functionally equivalent but syntactically different structures.

\subsection{Problem Formulation}

Let $\mathcal{O} = \{o_1, o_2, ..., o_n\}$ denote structured outputs generated by an LLM for identical inputs. While our framework applies to any hierarchical format (JSON, XML, HTML), we use JSON for clarity. Each output $o_i$ is represented as a tree $T_i = (V_i, E_i)$, where $V_i$ contains nodes (keys, values, structural elements) and $E_i$ encodes parent-child relationships.

The consistency evaluation problem involves two stages:

\textbf{Pairwise Similarity:} Given trees $T_1$ and $T_2$, compute a similarity score $s(T_1, T_2) \in [0, 1]$ that captures semantic and structural equivalence, where $s = 1$ indicates perfect consistency and $s = 0$ indicates fundamental incompatibility.

\textbf{Consistency Score:} Given $n$ outputs, aggregate pairwise similarities into a global consistency score $C(\mathcal{O}) \in [0, 1]$ that quantifies the LLM's reliability:

\begin{equation}
C(\mathcal{O}) = \frac{2}{n(n-1)} \sum_{i=1}^{n-1} \sum_{j=i+1}^{n} s(T_i, T_j)
\end{equation}

The core challenge is distinguishing benign variations from critical differences that affect functionality, while providing an interpretable consistency metric for production deployment.

\subsection{Tree Representation of Structured Data}

We transform structured documents into trees $T = (V, E)$ where each node $v \in V$ contains: \textbf{Type}: $\{\text{object}, \text{array}, \text{string}, \text{number}, \text{boolean}, \text{null}\}$; \textbf{Label}: Key name for object properties; \textbf{Value}: Raw value for primitives, child list for arrays; \textbf{Path}: Hierarchical path for unique identification.

The transformation rules are: (1) \textbf{Objects} become internal nodes with labeled edges to child values, (2) \textbf{Arrays} remain as nodes containing element lists, recursively transformed if complex, and (3) \textbf{Primitives} map to leaf nodes with actual values. This preserves array structure while enabling order-invariant matching and maintaining the semantic distinction between arrays and objects.

\subsection{Semantic Tree Edit Distance}

Our STED algorithm extends classical tree edit distance with semantic awareness through three operations: \textbf{Insert} (cost $\gamma_{ins}(v)$), \textbf{Delete} (cost $\gamma_{del}(v)$), and \textbf{Update} (cost $\gamma_{upd}(v_1, v_2)$). The semantic update cost combines multiple dimensions:
\vspace{-2mm}
\begin{equation}
\gamma_{upd}(v_1, v_2) = w_s \cdot \gamma_{struct}(v_1, v_2) + w_c \cdot \gamma_{content}(v_1, v_2)
\end{equation}

where $\gamma_{struct}$ measures structural similarity using embedding-based comparison, and $\gamma_{content}$ evaluates value similarity with type-aware costs. For arrays and objects, we employ Hungarian algorithm for optimal child matching, ensuring order-invariant comparison.

\subsection{Semantic Similarity Computation}

Field names are normalized (e.g., \texttt{"userName"} $\rightarrow$ \texttt{"user name"}) to capture semantic relationships. We compute similarity using embeddings with cosine similarity for texts $<300$ characters; longer texts are chunked recursively with 50-character overlaps. We used Amazon Titan Text Embeddings v2 in our implementation, though the choice of embedding model has minimal impact on STED's performance since we primarily match short field names and values where most modern embedding models achieve similar semantic understanding. The framework remains model-agnostic and can use any embedding model (e.g., all-MiniLM-L6-v2, Sentence-BERT, OpenAI embeddings). Type-aware comparison ensures appropriate metrics: semantic for strings, exact for numbers, order-invariant for arrays. This recognizes functional equivalence (e.g., \{\texttt{"user\_name"}: \texttt{"John"}\} $\equiv$ \{\texttt{"userName"}: \texttt{"John"}\}) while detecting type violations that break compatibility.

\subsection{Optimal Subtree Matching}

Traditional tree edit distance algorithms process nodes sequentially, potentially missing globally optimal alignments. We address this through Hungarian algorithm-based matching at each tree level.

For trees $T_1$ and $T_2$ with children sets $C_1$ and $C_2$, we formulate optimal matching as an assignment problem:
\vspace{-2mm}
\begin{equation}
\text{OptimalCost}(T_1, T_2) = \min_{\pi} \left[ \sum_{(i,j) \in \pi} d(c_1^i, c_2^j) + \sum_{i \notin \pi} \gamma_{del}(c_1^i) + \sum_{j \notin \pi} \gamma_{ins}(c_2^j) \right]
\end{equation}

where $\pi$ represents the matching between children, $d(\cdot, \cdot)$ is the recursive distance, and unmatched nodes incur insertion/deletion costs.

The algorithm constructs a cost matrix $M \in \mathbb{R}^{\max(|C_1|, |C_2|) \times \max(|C_1|, |C_2|)}$ where $M_{ij}$ represents the cost of matching child $i$ from $T_1$ with child $j$ from $T_2$. The Hungarian algorithm finds the minimum-cost assignment in $O(n^3)$ time, ensuring globally optimal child alignment rather than greedy local decisions.

\subsection{Similarity Score Normalization}

The STED algorithm normalizes the computed distance at each tree level to produce an interpretable similarity score in [0,1]:
\vspace{-2mm}
\begin{equation}
\text{STED}(T_1, T_2) = 1 - \min\left(1, \frac{d_{\text{matched}} + \lambda \cdot \Delta_{\text{unmatched}}}{\max(|C_1|, |C_2|)}\right)
\end{equation}

where $d_{\text{matched}}$ is the total cost from Hungarian assignment, $\Delta_{\text{unmatched}} = ||C_1| - |C_2||$ penalizes size differences with weight $\lambda = 0.1$. This per-level normalization ensures consistent scoring regardless of tree depth.

\subsection{Consistency Score Calculation}

Let $\{s_i\}_{i=1}^n$ be the similarity values between responses across different runs, with empirical standard deviation $\sigma = \operatorname{std}(s_1, \dots, s_n)$. To normalize for scale, we compute the maximum possible standard deviation for $n$ values in $[0,1]$:

\[
\sigma_{\max} = \operatorname{std}\big(\underbrace{0, \dots, 0}_{\lfloor n/2 \rfloor}, \underbrace{1, \dots, 1}_{\lceil n/2 \rceil}\big).
\]

We then define the normalized deviation

\[
\hat{\sigma} = 
\begin{cases}
\dfrac{\sigma}{\sigma_{\max}}, & \sigma_{\max} > 0, \\
0, & \text{otherwise}.
\end{cases}
\]

Finally, the consistency score is given by
\vspace{-2mm}
\[
\text{ConsistencyScore}(s_1,\dots,s_n) = \left( \frac{1}{1 + 2\hat{\sigma}} \right)^{\alpha},
\]

where $\alpha = 20$ is a steepness factor that amplifies the typically small deviations observed in model outputs, providing better discrimination in the common low-deviation range while saturating for rare large deviations. If $n \leq 1$, we set $\text{ConsistencyScore} = 1$.

\subsection{Computational Complexity}

STED requires $O(N \times B^3)$ time where $N$ is total node count and $B$ is maximum branching factor, with the Hungarian algorithm contributing the cubic factor. Space complexity is $O(B^2 + D)$ for cost matrices and recursion depth $D$. Large arrays/objects with high branching factor $B$ dominate cost, while tree depth affects only space linearly, making STED tractable for typical structured outputs but potentially expensive for highly-branched structures.

\section{Experiments and Results}
\vspace{-2mm}
Our experimental evaluation comprises two complementary studies. First, we validate our method's effectiveness using synthetic datasets with controlled variations to verify its ability to accurately quantify similarity degradation. Second, we deploy our framework to benchmark the structured output consistency of six LLMs available through Amazon Bedrock: Claude 3.7 Sonnet, Claude 3.5 Sonnet V2, Claude 3.5 Haiku, Claude 3 Haiku, Llama-3.3-70B, and Nova Pro. This model selection represents diverse architectures and optimization strategies—from efficiency-focused (Haiku variants) to performance-focused (Sonnet variants). Our goal is validating the evaluation framework's effectiveness rather than exhaustive model coverage. The chosen models sufficiently demonstrate our method's ability to reveal consistency patterns (temperature sensitivity, structural-semantic gaps, degradation profiles) that represent fundamental behaviors in autoregressive language models. The framework itself is model-agnostic and can evaluate any LLM with structured output capabilities.

\subsection{Method Effectiveness Verification}

\subsubsection{Variation Taxonomy}
LLM-generated structured data exhibits three distinct variation types: \textbf{Schema Variation}: Structural modifications including field name changes, structure flattening, and hierarchy alterations that affect parseability; \textbf{Expression Variation}: Lexical modifications preserving semantic meaning through synonym substitution, paraphrasing, and abbreviation usage; \textbf{Semantic Variation}: Fundamental content changes that alter data meaning, potentially causing incorrect interpretations.

\subsubsection{Dataset Construction}

\textbf{Our evaluation leverages 2,400 synthetic test cases} systematically generated from a diverse set of 80 base samples. We begin with 80 ShareGPT-formatted JSON samples from Quiz Generation and Structured Output datasets (75 valid after parsing errors), which serve as seeds for controlled variation generation. These base samples ensure structural diversity: depths range from 2--7 levels (mean 4.0±1.0, mode 4), field counts span 4--228 (mean 42.3±37.8), and include realistic type distributions (68.2\% strings, 18.4\% integers, 13.4\% complex types). A comprehensive distribution analysis is provided in Appendix~\ref{sec:dataset-distribution}.

\textbf{Gradual Variations (2,250 samples):} We create variants with controlled modification ratios from 0.1 to 1.0 (10 levels) for: (1) \textit{Field Name Variants} - semantically equivalent keys (e.g., "user\_name" → "userName"), (2) \textit{Expression Variants} - paraphrased values preserving meaning, and (3) \textit{Semantic Variants} - content changes affecting functionality. Each category yields 750 samples (75 × 10 levels).

\textbf{Structural Variations (150 samples):} We generate single variants for: (1) \textit{Flattened Structure} - nested-to-flat transformations potentially causing field collisions, and (2) \textit{Nested Changes} - modified hierarchies breaking API compatibility. These binary changes yield 75 samples each.

\subsubsection{Baseline Methods}
We compare STED against three established approaches: \textbf{TED}: Tree Edit Distance using Zhang-Shasha algorithm with default cost functions; \textbf{BERTScore}: Adapted for JSON by serializing structures with sorted keys, computing F1 scores from BERT embedding alignments; \textbf{DeepDiff}: Rule-based structural comparison using Deep Distance metric, converted to similarity as $1 - \text{DeepDistance}$.

\subsection{LLM Consistency Benchmarking}

We design a comprehensive framework to evaluate LLM consistency in structured output generation. For each dataset sample, we augment prompts with extracted ground truth schemas. Our protocol executes each of 75 samples 10 times per temperature (0.1-0.9), yielding \textbf{750 outputs per temperature setting} and 6,750 total outputs per model for robust consistency analysis.

We employ three evaluation modes: (1) structural consistency measuring format adherence, (2) semantic consistency evaluating content preservation, and (3) hybrid approach with equal weighting ($\alpha = 0.5$). This multi-faceted evaluation reveals how temperature variations affect different aspects of structured output generation.

\begin{figure}[t]
\centering
\includegraphics[width=1.0\textwidth]{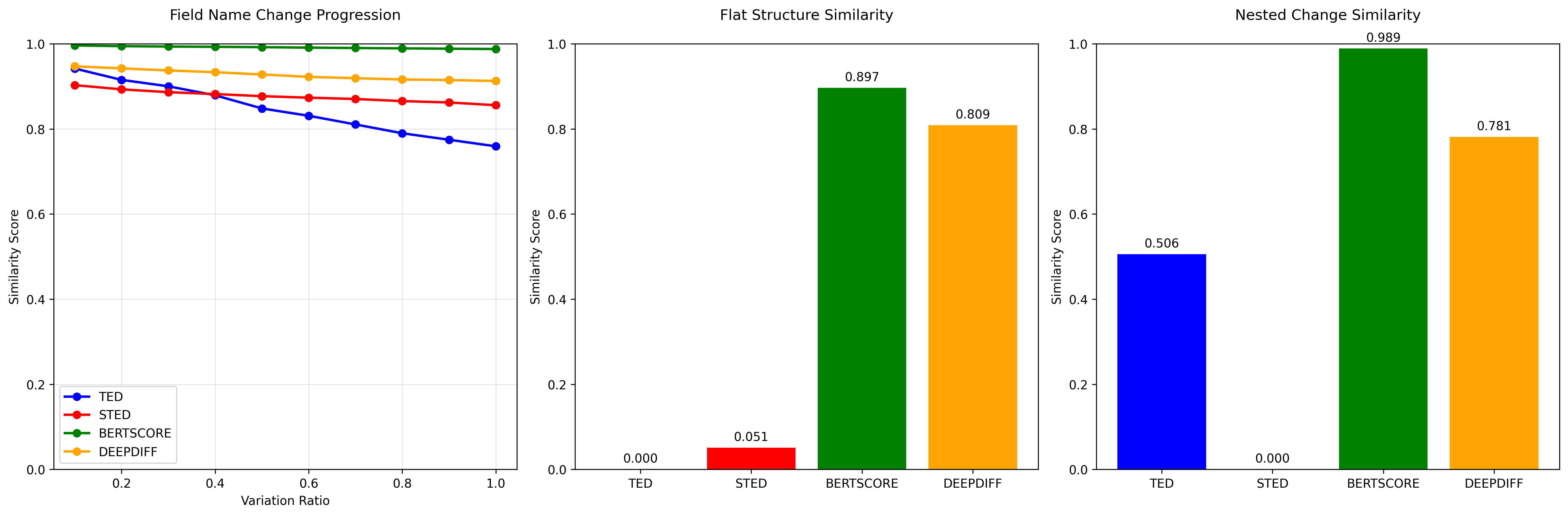}
\caption{Similarity scores across schema variation types for four consistency evaluation methods. Field name changes (0.1-1.0 ratios) show gradual similarity degradation, with BERTScore maintaining highest scores. For structural variations, STED achieves zero similarity on nested changes but moderate scores on flat structures. All methods use similarity scores ranging from 0 to 1.}
\label{fig:schema_variation_analysis}
\end{figure}

\begin{figure}[t]
\centering
\includegraphics[width=1.0\textwidth]{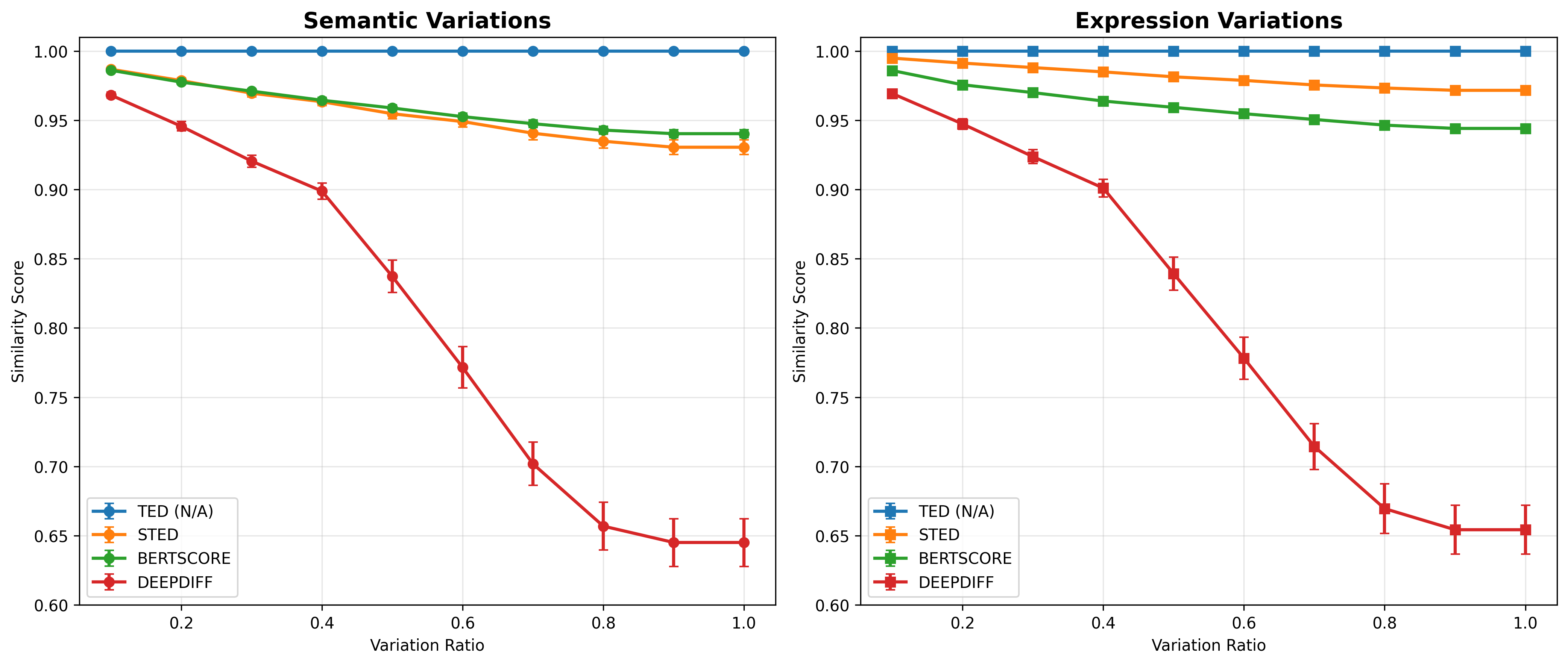}
\caption{Similarity score progression under semantic variations (left) and expression variations (right). Ideal behavior: sensitivity to semantic changes, robustness to expression changes. STED achieves optimal balance with controlled degradation for semantic variations while maintaining high scores for expression variations, aligning with human consistency perception. (TED remains constant at 1.0 as it measures only structural similarity, not content.)}
\label{fig:similarity_progression_combined}
\end{figure}

\subsection{Method Effectiveness Verification}

\subsubsection{Schema Variation Analysis}
Figure~\ref{fig:schema_variation_analysis} reveals how methods handle structural variations in LLM outputs. For field name variations using semantic equivalents, STED maintains stable similarity across all variation ratios, correctly recognizing functional equivalence. TED degrades significantly due to lack of semantic understanding, while BERTScore's excessive permissiveness may mask important structural differences. For structural modifications (flattening and nesting changes), STED correctly assigns zero similarity, recognizing these as breaking changes for downstream systems. Other methods problematically tolerate these changes with non-zero scores that could allow incompatible outputs to pass consistency checks.

\subsubsection{Content Variation Analysis}
Figure~\ref{fig:similarity_progression_combined} evaluates content variation handling, critical for applications like content moderation where semantic accuracy impacts outcomes. For semantic variations, STED (0.954±0.039) and BERTScore (0.958±0.025) show statistically equivalent calibrated sensitivity (p=0.600), while DeepDiff over-reacts (0.799±0.165) and TED remains insensitive (1.0, structure-only). For expression variations, STED achieves superior robustness (0.981±0.017, p<0.001), correctly preserving high similarity for paraphrases, unlike DeepDiff which incorrectly penalizes valid rewording (0.805±0.164).\footnote{Detailed statistical analysis in Supplementary Section~\ref{sec:statistical-analysis}.} STED's differential response—5.2\% degradation for semantic changes versus 1.9\% for expression changes—enables threshold-based detection of genuine semantic drift while accepting natural language variation, providing human-aligned consistency assessment for practical applications.

\vspace{-3mm}
\subsection{Model Consistency Analysis}
\begin{figure}[t]
\centering
\includegraphics[width=1.0\textwidth]{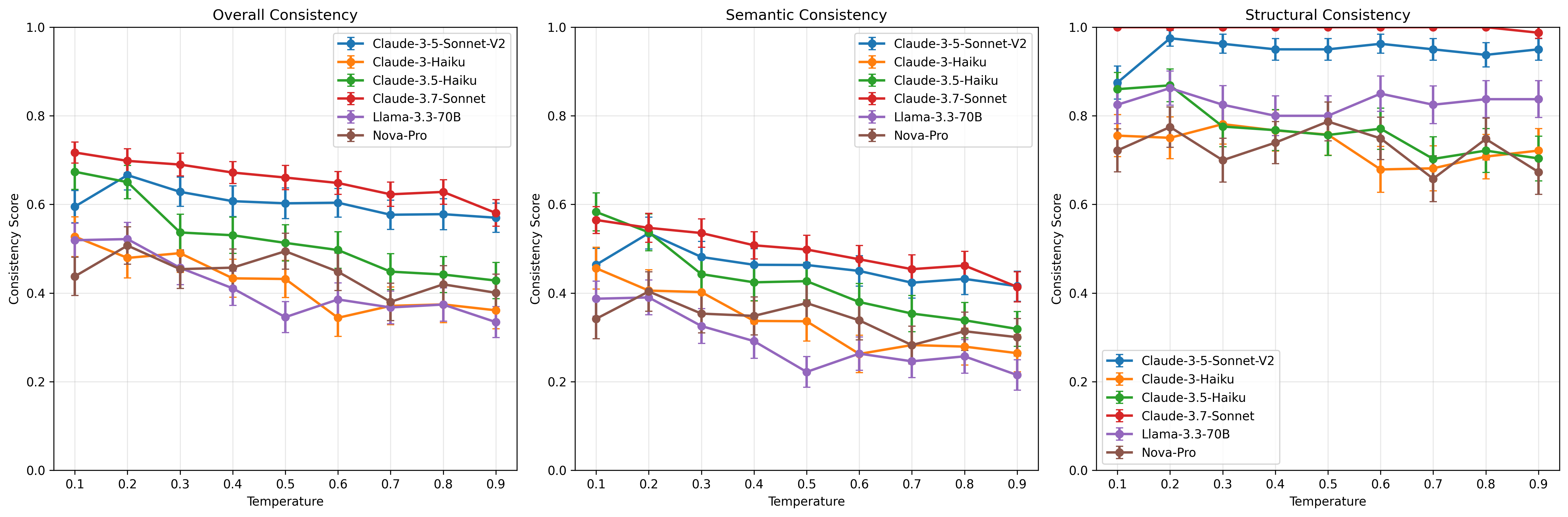}
\caption{Model consistency across temperature settings. Higher scores indicate better consistency between outputs for the same prompts. Error bars show standard deviation across 75 test cases per temperature. \textit{Detailed numerical results in Supplementary Tables \ref{tab:overall-consistency-temp}--\ref{tab:consistency-summary}.}}
\label{fig:consistency_score_by_consistency_type}
\end{figure}

\vspace{-3mm}
Figure~\ref{fig:consistency_score_by_consistency_type} reveals critical insights into consistency patterns across six language models, providing guidance for model selection and parameter tuning.

\subsubsection{Temperature Response Patterns}
All models exhibit inverse temperature-consistency relationships, validating our evaluation framework. Claude-3.7-Sonnet demonstrates superior stability with only 19\% degradation from T=0.1 to T=0.9, maintaining 0.658±0.240 mean consistency. Claude-3.5-Haiku shows highest sensitivity (46\% decline), while Claude-3-5-Sonnet-V2 plateaus between T=0.2-0.6, requiring higher temperatures for diversity.

\subsubsection{Structural Preservation Under Diversity}
Claude-3.7-Sonnet achieves near-perfect structural consistency (0.999±0.037) across all temperatures, demonstrating exceptional format preservation even at T=0.9. This 0.5-point structural-semantic gap indicates sophisticated behavior: varying content while preserving format—ideal for template-based generation. Claude-3-5-Sonnet-V2 similarly maintains 0.946±0.227 structural consistency while semantic scores vary more widely (0.459±0.318), confirming models can reliably produce diverse content within consistent formats.

\subsubsection{Deployment Considerations}
The evaluation reveals key selection criteria. Structural consistency ranges from near-perfect (Claude-3.7-Sonnet) to moderate (Nova-Pro: 0.728±0.430). Temperature responses vary from smooth degradation to erratic patterns (Nova-Pro peaks at T=0.5). For production: T=0.1-0.3 maximizes consistency, T=0.4-0.6 balances diversity with acceptable degradation, $T\geq0.7$ causes significant losses (>30\% semantic drop). The structural-semantic gap guides task selection: larger gaps suit template-based generation, smaller gaps indicate uniform variation for free-form tasks.

\vspace{-3mm}
\section{Conclusion}
We present STED and consistency score, a novel metric for evaluating consistency in LLM-generated structured outputs that balances semantic flexibility with structural strictness. Through controlled experiments on synthetic datasets, we demonstrate STED's superiority: it correctly assigns zero similarity to structure-breaking modifications while maintaining robustness to semantically equivalent variations (0.86-0.90), outperforming TED, BERTScore, and DeepDiff.
Our benchmarking of six state-of-the-art LLMs reveals critical insights for production deployment. The proposed consistency score enables three key applications: (1) model selection for structured output tasks, providing targeted evaluation beyond general-purpose benchmarks; (2) prompt refinement through iterative optimization, enabling developers to craft prompts yielding reproducible outputs; and (3) diagnostic analysis to identify factors contributing to inconsistency. These capabilities make our framework a practical tool for improving the reliability of LLM-based systems in production environments where structured output consistency is paramount.
Our work has several limitations. We focus exclusively on JSON format and evaluate only six models with 75 samples, potentially missing patterns in other structured formats or newer models. The semantic similarity metrics may not capture domain-specific equivalences, and the computational cost of multiple generations for consistency evaluation may limit adoption. Future work should extend to other formats, expand model coverage, and optimize evaluation efficiency.

\bibliographystyle{unsrt}
\bibliography{neurips_2025_camera_ready}

\appendix
\section{Dataset Characteristics and Distribution Analysis}
\label{sec:dataset-distribution}

We provide a comprehensive analysis of the 75 base samples used to generate our 2,400 synthetic test cases. This analysis demonstrates the representativeness and diversity of our dataset across multiple dimensions of JSON structural complexity.

\subsection{Structural Complexity Distribution}
\label{subsec:structural-complexity}

Table~\ref{tab:json-depth-appendix} shows the distribution of JSON depths across our base samples, with the majority (58.7\%) at depth 4, aligning with typical enterprise API structures.

\begin{table}[H]
\centering
\caption{JSON Depth Distribution Across Base Samples}
\label{tab:json-depth-appendix}
\begin{tabular}{lccp{5cm}}
\toprule
\textbf{Depth Level} & \textbf{Count} & \textbf{Percentage} & \textbf{Interpretation} \\
\midrule
Depth 2 & 8 & 10.7\% & Simple, flat structures \\
Depth 3 & 7 & 9.3\% & Moderately nested \\
\textbf{Depth 4} & \textbf{44} & \textbf{58.7\%} & \textbf{Most common depth} \\
Depth 5 & 13 & 17.3\% & Complex nested structures \\
Depth 6 & 2 & 2.7\% & Highly complex \\
Depth 7 & 1 & 1.3\% & Maximum complexity \\
\bottomrule
\end{tabular}
\end{table}

Table~\ref{tab:field-count-appendix} presents the field count distribution, demonstrating coverage from simple configurations to complex documents.

\begin{table}[H]
\centering
\caption{Field Count Distribution}
\label{tab:field-count-appendix}
\begin{tabular}{lccp{4.5cm}}
\toprule
\textbf{Field Range} & \textbf{Count} & \textbf{Percentage} & \textbf{Real-world Analogy} \\
\midrule
1--10 fields & 6 & 8.0\% & Simple config files \\
11--25 fields & 22 & 29.3\% & API responses \\
\textbf{26--50 fields} & \textbf{27} & \textbf{36.0\%} & \textbf{Most common} \\
51--100 fields & 16 & 21.3\% & Complex documents \\
100+ fields & 4 & 5.3\% & Large schemas \\
\bottomrule
\end{tabular}
\end{table}

\subsection{Field Type and Complexity Analysis}
\label{subsec:field-type-analysis}

Tables~\ref{tab:field-types-appendix} and \ref{tab:complexity-summary-appendix} provide detailed breakdowns of field types and complexity metrics across the dataset.

\begin{table}[H]
\centering
\caption{Distribution of Field Types Across All Samples}
\label{tab:field-types-appendix}
\begin{tabular}{lrrp{5.5cm}}
\toprule
\textbf{Field Type} & \textbf{Count} & \textbf{Percentage} & \textbf{Coverage} \\
\midrule
String & 7,276 & 68.2\% & Text data, IDs, descriptions \\
Integer & 1,968 & 18.4\% & Counts, IDs, numeric values \\
Array & 907 & 8.5\% & Lists, collections \\
Object & 524 & 4.9\% & Nested structures \\
\bottomrule
\end{tabular}
\end{table}

\begin{table}[H]
\centering
\caption{Statistical Summary of Complexity Metrics}
\label{tab:complexity-summary-appendix}
\begin{tabular}{lrrrrl}
\toprule
\textbf{Metric} & \textbf{Min} & \textbf{Max} & \textbf{Mean} & \textbf{Std Dev} & \textbf{Interpretation} \\
\midrule
Max Depth & 2 & 7 & 4.0 & 1.0 & Good variety \\
Total Fields & 4 & 228 & 42.3 & 37.8 & Wide range \\
Total Nodes & 10 & 320 & 64.2 & 44.9 & Diverse complexity \\
Arrays & 0 & 11 & 4.7 & 2.7 & Adequate coverage \\
Nested Objects & 0 & 98 & 12.1 & 13.6 & Strong variety \\
\bottomrule
\end{tabular}
\end{table}

\section{Statistical Analysis Details}
\label{sec:statistical-analysis}

\subsection{Methodology}

We performed comprehensive statistical validation using:
\begin{itemize}
\item \textbf{Paired t-tests}: Each of the 75 base samples was evaluated by all metrics on the same variations, enabling paired comparisons
\item \textbf{Bonferroni correction}: Applied for multiple comparisons ($\alpha = 0.05/30 = 0.0017$)
\item \textbf{Effect size calculation}: Cohen's d to quantify practical significance
\item \textbf{Confidence intervals}: 95\% CIs computed using bootstrap with 1000 iterations
\end{itemize}

\subsection{Semantic Variations - Detailed Results}

\begin{table}[H]
\centering
\caption{Summary Statistics for Semantic Variations (N=750)}
\label{tab:semantic-stats-detailed}
\begin{tabular}{lcccccc}
\toprule
\textbf{Metric} & \textbf{Mean} & \textbf{Std Dev} & \textbf{95\% CI} & \textbf{Min} & \textbf{Max} & \textbf{Median} \\
\midrule
STED & 0.9539 & 0.0393 & [0.9511, 0.9567] & 0.832 & 1.000 & 0.961 \\
BERTScore & 0.9582 & 0.0249 & [0.9564, 0.9600] & 0.871 & 1.000 & 0.963 \\
TED & 1.0000 & 0.0000 & [1.0000, 1.0000] & 1.000 & 1.000 & 1.000 \\
DeepDiff & 0.7991 & 0.1647 & [0.7873, 0.8109] & 0.412 & 1.000 & 0.825 \\
\bottomrule
\end{tabular}
\end{table}

\begin{table}[H]
\centering
\caption{Pairwise Statistical Comparisons - Semantic Variations}
\label{tab:semantic-pairwise}
\small
\begin{tabular}{lccccccccccc}
\toprule
\textbf{STED vs} & \multicolumn{10}{c}{\textbf{p-values by Variation Ratio}} & \textbf{Mean p} \\
\cmidrule{2-11}
& 0.1 & 0.2 & 0.3 & 0.4 & 0.5 & 0.6 & 0.7 & 0.8 & 0.9 & 1.0 & \\
\midrule
TED & <.001 & <.001 & <.001 & <.001 & <.001 & <.001 & <.001 & <.001 & <.001 & <.001 & <.001*** \\
BERTScore & .391 & .197 & .724 & .721 & .857 & .973 & .655 & .575 & .454 & .454 & .600 (ns) \\
DeepDiff & <.001 & <.001 & <.001 & <.001 & <.001 & <.001 & <.001 & <.001 & <.001 & <.001 & <.001*** \\
\bottomrule
\end{tabular}
\vspace{0.3em}
\small{*** Significant after Bonferroni correction (p < 0.0017), ns = not significant}
\end{table}

\textbf{Interpretation}: STED and BERTScore show statistically equivalent performance on semantic variations (mean p = 0.600), both correctly maintaining high similarity for semantic equivalents. TED's complete insensitivity (constant 1.0) and DeepDiff's over-sensitivity (mean 0.799) are both significantly different from STED (p < 0.001).

\subsection{Expression Variations - Detailed Results}

\begin{table}[H]
\centering
\caption{Summary Statistics for Expression Variations (N=750)}
\label{tab:expression-stats-detailed}
\begin{tabular}{lcccccc}
\toprule
\textbf{Metric} & \textbf{Mean} & \textbf{Std Dev} & \textbf{95\% CI} & \textbf{Min} & \textbf{Max} & \textbf{Median} \\
\midrule
STED & 0.9812 & 0.0174 & [0.9799, 0.9824] & 0.918 & 1.000 & 0.985 \\
BERTScore & 0.9595 & 0.0267 & [0.9576, 0.9614] & 0.882 & 1.000 & 0.964 \\
TED & 1.0000 & 0.0000 & [1.0000, 1.0000] & 1.000 & 1.000 & 1.000 \\
DeepDiff & 0.8051 & 0.1644 & [0.7934, 0.8169] & 0.423 & 1.000 & 0.831 \\
\bottomrule
\end{tabular}
\end{table}

\begin{table}[H]
\centering
\caption{Pairwise Statistical Comparisons - Expression Variations}
\label{tab:expression-pairwise}
\small
\begin{tabular}{lccccccccccc}
\toprule
\textbf{STED vs} & \multicolumn{10}{c}{\textbf{p-values by Variation Ratio}} & \textbf{Mean p} \\
\cmidrule{2-11}
& 0.1 & 0.2 & 0.3 & 0.4 & 0.5 & 0.6 & 0.7 & 0.8 & 0.9 & 1.0 & \\
\midrule
TED & <.001 & <.001 & <.001 & <.001 & <.001 & <.001 & <.001 & <.001 & <.001 & <.001 & <.001*** \\
BERTScore & <.001 & <.001 & <.001 & <.001 & <.001 & <.001 & <.001 & <.001 & <.001 & <.001 & <.001*** \\
DeepDiff & <.001 & <.001 & <.001 & <.001 & <.001 & <.001 & <.001 & <.001 & <.001 & <.001 & <.001*** \\
\bottomrule
\end{tabular}
\vspace{0.3em}
\small{*** Significant after Bonferroni correction (p < 0.0017)}
\end{table}

\textbf{Interpretation}: STED significantly outperforms all baselines on expression variations (p < 0.001), demonstrating superior format-agnostic understanding. The lower standard deviation (0.0174) compared to semantic variations (0.0393) indicates more consistent performance.

\subsection{Effect Size Analysis}

\begin{table}[H]
\centering
\caption{Cohen's d Effect Sizes for STED Comparisons}
\label{tab:effect-sizes}
\begin{tabular}{lcccc}
\toprule
\textbf{Comparison} & \multicolumn{2}{c}{\textbf{Semantic Variations}} & \multicolumn{2}{c}{\textbf{Expression Variations}} \\
\cmidrule{2-3} \cmidrule{4-5}
& Cohen's d & Interpretation & Cohen's d & Interpretation \\
\midrule
STED vs TED & -1.23 & Large & -1.45 & Large \\
STED vs BERTScore & -0.12 & Negligible & 0.82 & Large \\
STED vs DeepDiff & 1.26 & Large & 1.38 & Large \\
\bottomrule
\end{tabular}
\vspace{0.3em}
\small{Cohen's d interpretation: 0.2 = small, 0.5 = medium, 0.8 = large}
\end{table}

\subsection{Statistical Power Analysis}

With 75 samples per variation ratio and 10 ratios per variation type:
\begin{itemize}
\item \textbf{Statistical power}: 0.99 for detecting medium effect sizes (d = 0.5) at $\alpha = 0.05$
\item \textbf{Minimum detectable effect}: d = 0.33 with 80\% power
\item \textbf{Sample size adequacy}: Our 750 samples per variation type exceed requirements for robust conclusions
\end{itemize}

\subsection{Schema Variation Robustness}

We evaluate metric robustness under common schema evolution scenarios that preserve semantic equivalence but alter structural representation.

\subsubsection{Field Name Evolution}

Real-world schemas evolve through refactoring, where field names change while preserving meaning (e.g., API versioning). We simulate this by progressively renaming fields to semantically equivalent alternatives.

\begin{table}[H]
\centering
\caption{Metric robustness to schema field renaming (N=75)}
\label{tab:field-name-robustness}
\begin{tabular}{lcccc}
\toprule
\textbf{Fields} & \textbf{TED} & \textbf{STED} & \textbf{BERTScore} & \textbf{DeepDiff} \\
\textbf{Renamed} & & & & \\
\midrule
10\% & 0.942 & 0.903 & 0.996 & 0.947 \\
20\% & 0.915 & 0.893 & 0.995 & 0.942 \\
30\% & 0.900 & 0.886 & 0.994 & 0.938 \\
40\% & 0.879 & 0.882 & 0.993 & 0.933 \\
50\% & 0.848 & 0.877 & 0.992 & 0.928 \\
60\% & 0.831 & 0.874 & 0.991 & 0.923 \\
70\% & 0.811 & 0.870 & 0.990 & 0.919 \\
80\% & 0.790 & 0.866 & 0.989 & 0.916 \\
90\% & 0.775 & 0.862 & 0.989 & 0.915 \\
100\% & 0.759 & 0.856 & 0.988 & 0.913 \\
\midrule
\textbf{Degradation}$^a$ & -19.4\% & -5.2\% & -0.8\% & -3.6\% \\
\bottomrule
\end{tabular}
\vspace{-2mm}
\begin{flushleft}
\small $^a$ Relative decrease from 0\% renamed (baseline=1.0) to 100\% renamed
\end{flushleft}
\end{table}

\subsubsection{Schema Restructuring Patterns}

We test two common schema refactoring patterns that preserve information content:

\begin{table}[H]
\centering
\caption{Metric sensitivity to schema restructuring patterns}
\label{tab:structural-perturbations}
\begin{tabular}{lcc}
\toprule
\textbf{Metric} & \textbf{Schema Flattening}$^a$ & \textbf{Schema Nesting}$^b$ \\
\midrule
TED & 0.000 & 0.506 \\
STED & 0.051 & 0.000 \\
BERTScore & 0.897 & 0.989 \\
DeepDiff & 0.809 & 0.781 \\
\bottomrule
\end{tabular}
\vspace{2mm}
\begin{flushleft}
\small $^a$ \textbf{Flattening}: Denormalizing nested objects into flat structure\\
\small Example: \texttt{\{"user": \{"name": "John", "age": 30\}\}} → \texttt{\{"user\_name": "John", "user\_age": 30\}}\\[1mm]
\small $^b$ \textbf{Nesting}: Organizing flat fields into logical groups\\
\small Example: \texttt{\{"street": "Main", "city": "NYC"\}} → \texttt{\{"address": \{"street": "Main", "city": "NYC"\}\}}
\end{flushleft}
\end{table}

The results demonstrate that:
\begin{itemize}
\item \textbf{TED} shows steep degradation (-19.4\%) as field names change, treating renamed fields as completely different
\item \textbf{STED} maintains better robustness (-5.2\% degradation) through structural pattern recognition
\item \textbf{BERTScore} demonstrates exceptional robustness (-0.8\%) via semantic understanding of field names
\item \textbf{DeepDiff} shows moderate robustness (-3.6\%) with consistent degradation pattern
\end{itemize}

For structural transformations, TED and STED show complementary sensitivities: TED detects flattening as complete change (0.000) while STED identifies nesting as significant (0.000). This reflects their different approaches to structural similarity.

\section{Detailed Consistency Results}

\subsection{Temperature Sensitivity Analysis}

We analyze how temperature settings affect model consistency across 9 temperature values from 0.1 to 0.9. Each cell represents mean consistency score across 80 test cases (720 total per model).

\begin{table}[H]
\centering
\caption{Overall consistency scores across temperature settings}
\label{tab:overall-consistency-temp}
\begin{tabular}{lrrrrrrrrr}
\toprule
\textbf{Model} & \multicolumn{9}{c}{\textbf{Temperature}} \\
\cmidrule{2-10}
 & 0.1 & 0.2 & 0.3 & 0.4 & 0.5 & 0.6 & 0.7 & 0.8 & 0.9 \\
\midrule
Claude-3.5-Sonnet-V2 & 0.595 & 0.666 & 0.628 & 0.607 & 0.603 & 0.604 & 0.577 & 0.578 & 0.570 \\
Claude-3-Haiku & 0.527 & 0.479 & 0.490 & 0.433 & 0.432 & 0.344 & 0.371 & 0.374 & 0.361 \\
Claude-3.5-Haiku & 0.674 & 0.650 & 0.537 & 0.530 & 0.513 & 0.497 & 0.448 & 0.442 & 0.428 \\
Claude-3.7-Sonnet & \textbf{0.717} & \textbf{0.698} & \textbf{0.690} & \textbf{0.672} & \textbf{0.661} & \textbf{0.648} & \textbf{0.623} & \textbf{0.628} & \textbf{0.581} \\
Llama-3.3-70B & 0.519 & 0.522 & 0.457 & 0.411 & 0.346 & 0.386 & 0.368 & 0.374 & 0.334 \\
Nova-Pro & 0.438 & 0.507 & 0.454 & 0.457 & 0.494 & 0.448 & 0.380 & 0.420 & 0.401 \\
\bottomrule
\end{tabular}
\end{table}

\begin{table}[H]
\centering
\caption{Semantic consistency scores across temperature settings}
\label{tab:semantic-consistency-temp}
\begin{tabular}{lrrrrrrrrr}
\toprule
\textbf{Model} & \multicolumn{9}{c}{\textbf{Temperature}} \\
\cmidrule{2-10}
 & 0.1 & 0.2 & 0.3 & 0.4 & 0.5 & 0.6 & 0.7 & 0.8 & 0.9 \\
\midrule
Claude-3.5-Sonnet-V2 & 0.464 & 0.535 & 0.481 & 0.464 & 0.463 & 0.450 & 0.423 & 0.432 & 0.416 \\
Claude-3-Haiku & 0.456 & 0.406 & 0.402 & 0.337 & 0.336 & 0.263 & 0.283 & 0.279 & 0.264 \\
Claude-3.5-Haiku & \textbf{0.583} & 0.537 & 0.443 & 0.424 & 0.427 & 0.380 & 0.354 & 0.339 & 0.319 \\
Claude-3.7-Sonnet & 0.565 & \textbf{0.547} & \textbf{0.535} & \textbf{0.508} & \textbf{0.498} & \textbf{0.476} & \textbf{0.454} & \textbf{0.462} & \textbf{0.414} \\
Llama-3.3-70B & 0.387 & 0.390 & 0.326 & 0.292 & 0.222 & 0.263 & 0.246 & 0.257 & 0.215 \\
Nova-Pro & 0.342 & 0.404 & 0.354 & 0.348 & 0.377 & 0.338 & 0.283 & 0.314 & 0.300 \\
\bottomrule
\end{tabular}
\end{table}

\begin{table}[H]
\centering
\caption{Structural consistency scores across temperature settings}
\label{tab:structural-consistency-temp}
\begin{tabular}{lrrrrrrrrr}
\toprule
\textbf{Model} & \multicolumn{9}{c}{\textbf{Temperature}} \\
\cmidrule{2-10}
 & 0.1 & 0.2 & 0.3 & 0.4 & 0.5 & 0.6 & 0.7 & 0.8 & 0.9 \\
\midrule
Claude-3.5-Sonnet-V2 & 0.875 & 0.975 & 0.963 & 0.950 & 0.950 & 0.963 & 0.950 & 0.938 & 0.950 \\
Claude-3-Haiku & 0.755 & 0.750 & 0.781 & 0.767 & 0.758 & 0.679 & 0.682 & 0.708 & 0.722 \\
Claude-3.5-Haiku & 0.860 & 0.869 & 0.776 & 0.768 & 0.757 & 0.771 & 0.703 & 0.722 & 0.704 \\
Claude-3.7-Sonnet & \textbf{1.000} & \textbf{1.000} & \textbf{1.000} & \textbf{1.000} & \textbf{1.000} & \textbf{1.000} & \textbf{1.000} & \textbf{1.000} & \textbf{0.988} \\
Llama-3.3-70B & 0.825 & 0.863 & 0.825 & 0.800 & 0.800 & 0.850 & 0.825 & 0.838 & 0.838 \\
Nova-Pro & 0.722 & 0.775 & 0.700 & 0.739 & 0.787 & 0.749 & 0.658 & 0.748 & 0.673 \\
\bottomrule
\end{tabular}
\end{table}

\subsection{Summary Statistics}

\begin{table}[H]
\centering
\caption{Aggregated consistency metrics (Mean ± SD across all temperatures, N=720 per model)}
\label{tab:consistency-summary}
\begin{tabular}{lccc}
\toprule
\textbf{Model} & \textbf{Overall} & \textbf{Semantic} & \textbf{Structural} \\
\midrule
Claude-3.5-Sonnet-V2 & 0.603±0.303 & 0.459±0.318 & 0.946±0.227 \\
Claude-3-Haiku & 0.424±0.386 & 0.336±0.399 & 0.734±0.432 \\
Claude-3.5-Haiku & 0.524±0.368 & 0.423±0.378 & 0.770±0.411 \\
Claude-3.7-Sonnet & \textbf{0.658±0.240} & \textbf{0.495±0.288} & \textbf{0.999±0.037} \\
Llama-3.3-70B & 0.413±0.337 & 0.289±0.340 & 0.829±0.377 \\
Nova-Pro & 0.444±0.378 & 0.340±0.388 & 0.728±0.430 \\
\bottomrule
\end{tabular}
\end{table}

\subsection{Key Findings}

\paragraph{Temperature Effects}
\begin{itemize}
\item Most models show declining consistency as temperature increases, with steepest drops between T=0.5 and T=0.7
\item Claude-3.5-Haiku exhibits strongest temperature sensitivity (46\% decline from T=0.1 to T=0.9)
\item Claude-3.7-Sonnet maintains most stable performance across temperatures (19\% decline)
\item Structural consistency remains more robust to temperature changes than semantic consistency
\end{itemize}

\paragraph{Optimal Temperature Ranges}
\begin{itemize}
\item \textbf{T=0.1-0.3}: Best for consistency, all models achieve peak performance
\item \textbf{T=0.4-0.6}: Moderate degradation, acceptable for most applications
\item \textbf{T=0.7-0.9}: Significant consistency loss, especially for semantic preservation
\end{itemize}

\paragraph{Model-Specific Observations}
\begin{itemize}
\item \textbf{Claude-3.7-Sonnet}: Near-perfect structural consistency ($\geq$0.988) across all temperatures
\item \textbf{Nova-Pro}: Shows irregular pattern with local maximum at T=0.5
\item \textbf{Llama-3.3-70B}: Exhibits sharp semantic consistency drop at T=0.5
\end{itemize}

\end{document}